\documentclass[Afour,sageh,times,review]{sagej}

\usepackage{subfiles} 
\usepackage{moreverb,url}
\usepackage{xparse}
\usepackage[linesnumbered,ruled,vlined]{algorithm2e}
\usepackage{algpseudocode}
\usepackage{multirow}

\newcommand{\ie}{\emph{i.e.}, }
\newcommand{\eg}{\emph{e.g.}, }

\usepackage[colorlinks,bookmarksopen,bookmarksnumbered,citecolor=darkblue]{hyperref}
\usepackage{xcolor}
\definecolor{darkblue}{RGB}{0,0,139}
 
\newcommand\BibTeX{{\rmfamily B\kern-.05em \textsc{i\kern-.025em b}\kern-.08em
T\kern-.1667em\lower.7ex\hbox{E}\kern-.125emX}}

\def\volumeyear{2016}

\begin{document}

\runninghead{Jiaming Chen et al.}


\title{Vision-Language Model Predictive Control for Manipulation Planning and Trajectory Generation}

\author{Jiaming Chen$^*$\affilnum{1,2}, Wentao Zhao$^*$\affilnum{1}, Ziyu Meng\affilnum{1}, Donghui Mao\affilnum{1}, Ran Song\affilnum{1}, Wei Pan\affilnum{2}, Wei Zhang\affilnum{1}}

\affiliation{\scriptsize\affilnum{1} School of Control Science and Engineering, Shandong University, China\\
\affilnum{2} Department of Computer Science, The University of Manchester, UK}

\corrauth{Ran Song, School of Control Science and Engineering,
Shandong University,
Jinan, 250061, China.}

\email{ransong@sdu.edu.cn}

\begin{abstract}
Model Predictive Control (MPC) is a widely adopted control paradigm that leverages predictive models to estimate future system states and optimize control inputs accordingly. However, while MPC excels in planning and control, it lacks the capability for environmental perception, leading to failures in complex and unstructured scenarios.
To address this limitation, we introduce Vision-Language Model Predictive Control (VLMPC), a robotic manipulation planning framework that integrates the perception power of vision-language models (VLMs) with MPC. VLMPC utilizes a conditional action sampling module that takes a goal image or language instruction as input and leverages VLM to generate candidate action sequences. These candidates are fed into a video prediction model that simulates future frames based on the actions. In addition, we propose an enhanced variant, Traj-VLMPC, which replaces video prediction with motion trajectory generation to reduce computational complexity while maintaining accuracy. Traj-VLMPC estimates motion dynamics conditioned on the candidate actions, offering a more efficient alternative for long-horizon tasks and real-time applications.
Both VLMPC and Traj-VLMPC select the optimal action sequence using a VLM-based hierarchical cost function that captures both pixel-level and knowledge-level consistency between the current observation and the task input. We demonstrate that both approaches outperform existing state-of-the-art methods on public benchmarks and achieve excellent performance in various real-world robotic manipulation tasks.
Code is available at~\href{https://github.com/PPjmchen/VLMPC}{https://github.com/PPjmchen/VLMPC}.
\end{abstract}

\maketitle

\footnotetext[1]{These authors contributed equally to this work.}

\section{Introduction}

Burgeoning foundation models~\citep{gpt4,gpt3, palm,bommasani2021opportunities,palm-e} have demonstrated powerful capabilities of knowledge extraction and reasoning. Exploration based on foundation models has thus flourished in many fields such as computer vision~\citep{liu2023visual,chen2023minigpt,dai2023instructblip,bai2023sequential}, AI for science~\citep{bi2023accurate}, healthcare~\citep{moor2023foundation,thirunavukarasu2023large,zhou2023foundation,qiu2023large}, and robotics~\citep{rt-2,ha2023scaling,ren2023robots,yu2023language,mandi2023roco}. Recently, a wealth of work has made significant progress in incorporating foundation models into robotics. These works usually leveraged the strong understanding and reasoning capabilities of versatile foundation models on multimodal data, including language~\citep{voxposer,rt-2,ren2023robots,yu2023language,languagempc,mandi2023roco}, image~\citep{voxposer,liu2023aligning}, and video~\citep{rt-2}, to enhance robotic perception and decision making. 

To achieve knowledge transfer from foundation models to robots, most early works concentrate on task planning~\citep{huang2022language,huang2022inner,chen2023open,wang2023voyager,singh2023progprompt,raman2022planning,song2023llm,liu2023llm+,lin2023text2motion,ding2023task,yuan2023plan4mc,xie2023translating,lu2023multimodal,pallagani2024prospects,ni2023grid}, which directly utilize large language models (LLMs) to decompose high-level natural language command and abstract tasks into low-level and pre-defined primitives (\textit{i.e.,} executable actions or skills). Although such schemes intuitively enable robots to perform complex and long-horizon tasks, they lack the capability of visual perception. Consequently, they heavily rely on pre-defined individual skills to interact with specific physical entities, which limits the flexibility and applicability of robotic planning. Recent works~\citep{voxposer,rt-2,wake2023gpt,hu2023look} remedy this issue by integrating with large-scale vision-language models (VLMs) to improve scene perception and generate trajectories adaptively for robotic manipulation in intricate scenarios without using pre-defined primitives. 

Although existing methods have shown promising results in incorporating foundation models into robotic manipulation, interaction with a wide variety of objects and humans in the real world remains a challenge. Specifically, since the future states of a robot are not fully considered in the decision-making loop of such methods, the reasoning of foundation models is primarily based on current observations, resulting in insufficient forward-looking planning. For example, in the task of opening a drawer, the latest method based on VLM~\citep{voxposer} cannot directly generate an accurate trajectory to pull the drawer handle due to the lack of prediction on the future state, and thus it still requires designing specific primitives on object-level interaction. Hence, it is desirable to develop a robotic framework that performs with a human-like \emph{``look before you leap''} ability.

Model predictive control (MPC) is a control strategy widely used in robotics~\citep{shim2003decentralized,allibert2010predictive,howard2010receding,williams2017information,lenz2015deepmpc}. MPC possesses the appealing attribute of predicting the future states of a system through a predictive model. This forward-looking attribute allows robots to plan their actions by considering potential future scenarios, thus enhancing their ability to interact dynamically with various environments. Traditional MPC~\citep{shim2003decentralized,howard2010receding,williams2017information,mpc_robot_exp2,mpc_robot_exp3} usually builds a deterministic and sophisticated dynamic model corresponding to the task and environment, which does not adapt well to intricate scenes in the real world.  Recent research \citep{visual_foresight,ye2020object,nair2022learning,3dmpc,vp2,ebert2018robustness} has explored using vision-based predictive models to learn dynamic models from visual inputs and predict high-dimensional future states in 2D~\citep{visual_foresight,ye2020object,vp2,ebert2018robustness} or 3D~\citep{visual_foresight,nair2022learning,3dmpc,ebert2018robustness} spaces. Such methods are based on current visual observations for proposing manipulation plans in the MPC loop, which enables robots to make more reasonable decisions based on visual clues. However, the effectiveness of such methods is constrained by the limitations inherent in visual predictive models trained on finite datasets. Such models struggle to accurately predict scenarios involving scenes or objects they have not previously encountered. This issue becomes especially pronounced in the real-world environments, often partially or even fully unseen to robots, where the models can only perform basic tasks that align closely with their training data.

Naturally, large-scale VLMs have the potential to address this problem by providing extensive open-domain knowledge and offering a more comprehensive understanding of diverse and unseen scenarios, thereby enhancing the predictive accuracy and adaptability of the scheme for robotic manipulation. Thus, this work presents \textbf{V}ision-\textbf{L}anguage \textbf{M}odel \textbf{P}redictive \textbf{C}ontrol (\textbf{VLMPC}), a framework that combines VLM and model predictive control to guide robotic manipulation with complicated path planning including rotation and interaction with scene objects. By leveraging the strong ability of visual reasoning and visual grounding for sampling actions provided by VLM, VLMPC avoids the manual design of individual primitives, and addresses the limitation that previous methods based on VLMs can only compose coarse trajectories without foresight.

\begin{figure*}[t]
    \centering
    \includegraphics[height=9.3cm]{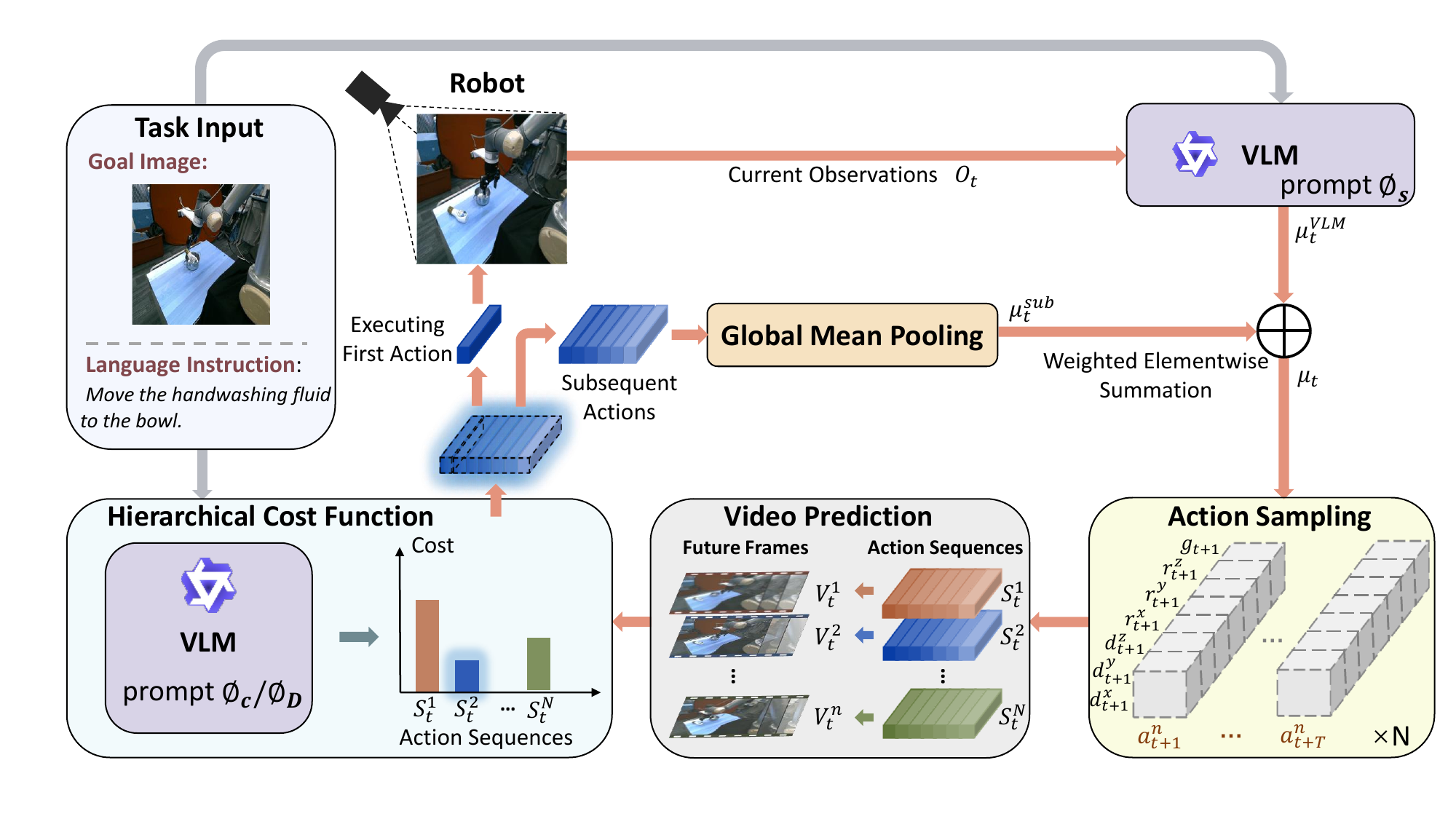}
    \caption{VLMPC takes as input either a goal image or a language instruction. It first prompts VLMs to generate a conditional sampling distribution, from which action sequences are derived. Then, such action sequences are fed into a lightweight action-conditioned video prediction model to predict a set of future frames. The assessment of VLMPC is performed with a hierarchical cost function composed of two sub-costs: a pixel distance cost and a VLM-assisted cost for performing video assessments based on the future frames. VLMPC finally selects the best action sequence, in which the robot picks the first action to execute and the subsequent actions are fed into the action sampling module to further assist conditional action sampling.}
    \label{fig1}
\end{figure*}

As illustrated in Fig.~\ref{fig1}, VLMPC takes as input either a goal image indicating the prospective state or a language instruction. We propose an action sampling module that uses VLM to initialize the task and handle the current observation, which generates a conditional action sampling distribution for further producing a set of action sequences. With the action sequences and the history image observation, VLMPC adopts a lightweight action-conditioned video prediction model to predict a set of future frames. To assess the quality of the candidate action sequences through the future frames, we also design a hierarchical cost function composed of two sub-costs: a pixel-level cost measuring the difference between the video predictions and the goal image and a knowledge-level cost making a comprehensive evaluation on the video predictions. VLMPC finally chooses the action sequence corresponding to the best video prediction, and then picks the first action from the sequence to execute while feeding the subsequent actions into the action sampling module combined with conditional action sampling. 

Compared to directly sampling and predicting within the executable action space, object trajectory provides a more efficient and stable solution for manipulation planning~\citep{bharadhwaj2024track2act,wen2023anypoint,xu2024flow,yuan2024general}. Trajectory-based methods leverage 2D~\citep{bharadhwaj2024track2act,wen2023anypoint,xu2024flow} or 3D~\citep{yuan2024general} observational inputs to predict the motion trajectory of the interacting objects or the robot. These methods offer several advantages: (1) they capture continuous and smooth trajectories, enhancing execution stability; (2) they enable more precise coordination between the robot and the environment; (3) they reduce the computational complexity associated with discrete action sampling. Additionally, the availability of large-scale robotic manipulation datasets~\citep{padalkar2023open}, rich in physical interaction data, has significantly advanced the development of models with inherent scene understanding and motion prediction capabilities, empowering robots to better model their surroundings and generate robust manipulation strategies.

Therefore, this work proposes an enhanced version of VLMPC that leverages motion trajectories, termed Traj-VLMPC (Trajectory-based Vision-Language Model Predictive Control). We design a VLM-driven Gaussian Mixture Model (GMM) to replace action sampling and video prediction by generating diverse and adaptive motion trajectories from the mixture of Gaussian distributions in 3D space, where VLM conditions the parameters of GMM. To assess the quality of the trajectories, we first generate a voxel-based 3D value map that assigns a contextual relevance score to each spatial position, reflecting its importance for achieving the desired task objectives, while also taking into account the task instructions, the target objects, and potential obstacles. Then, we propose a cost function that sums the waypoints' values of trajectories in the value map for trajectory assessment. Similar to VLMPC, the highest-ranked trajectory is adopted for the robot's action execution.


The main contributions of this paper are as follows:
\begin{enumerate}
    \item We propose VLMPC for robotic manipulation planning, which incorporates a learning-based dynamic model to predict future video frames and seamlessly integrates VLM into the MPC loop for open-set knowledge reasoning. 
    
    \item We design a conditional action sampling module to sample robot actions from a visual perspective and a hierarchical cost function to provide a comprehensive and coarse-to-fine assessment of video predictions.
    

    \item We introduce Traj-VLMPC, an enhanced variant of VLMPC, which incorporates a VLM-conditioned GMM as a 3D motion trajectory sampler and a generator and uses a VLM-based 3D value map for efficient trajectory evaluation.
    
    \item Experiments in simulated and real-world scenarios demonstrate that VLMPC and Traj-VLMPC achieve state-of-the-art performance without pre-defined primitives, where Traj-VLMPC significantly enhances control stability and execution speed in long-horizon tasks.
\end{enumerate}

This paper is an extended version of~\cite{vlmpc}, and contribution (3) is the main extension. The outline of the paper is as follows. In Sec.~\ref{sec:related_work}, we list and analyze related work, including MPC and foundation models for robotic manipulation. Sec.~\ref{sec:methods} introduces the proposed VLMPC framework and demonstrates each module in detail. In Sec.~\ref{sec:traj}, we further propose the enhanced variant, Traj-VLMPC. In Sec.~\ref{sec:experiments}, experiments on both simulated and real-world environments are carried out to validate the effectiveness of VLMPC and Traj-VLMPC. We summarize our work in Sec.~\ref{sec:conclusion}.

\section{Related Work}
\label{sec:related_work}

\subsection{Model Predictive Control for Robotic Manipulation}
Model predictive control (MPC) is a multivariate control algorithm widely used in robotics \citep{shim2003decentralized,allibert2010predictive,howard2010receding,williams2017information,lenz2015deepmpc,mpc-policy,mpc_robot_exp1,mpc_robot_exp2,mpc_robot_exp3,voxposer,visual_foresight}. It employs a predictive model to estimate future system states, subsequently formulating the control law through solving a constrained optimization problem~\citep{mpc_review,mpc-policy}. The foresight capability of MPC, combined with its constraint-handling features, enables the development of advanced robotic systems which operate safely and efficiently in variable environments~\citep{howard2010receding}.

In the context of robotic manipulation, the role of MPC is to make the robot manipulator move and act in an optimal way with respect to input and output constraints~\citep{bhardwaj2022storm,visual_foresight,deep_visual_foresight,3dmpc,ye2020object,nair2022learning,vp2}.
In particular, action-based predictive models are frequently used in MPC for robotic manipulation, referring to a prediction model designed to forecast the potential future outcomes of specific actions, which connect sequence data to decision-making processes.~\cite{bhardwaj2022storm} proposed a sampling-based MPC integrated with low discrepancy sampling, smooth trajectory generation, and behavior-based cost functions to produce good robot actions that reach the goal poses.
Visual Foresight ~\citep{visual_foresight,deep_visual_foresight} first generated robotic planning towards a specific goal by using a video prediction model to simulate candidate action sequences and then scored them based on the similarity between their predicted futures and the goal.~\cite{3dmpc} proposed a 3D volumetric scene representation that simultaneously discovers, tracks, and reconstructs objects and predicts their motion under the interactions of a robot.~\cite{ye2020object} presented an approach to learn an object-centric forward model, which planned action sequences to achieve distant desired goals. Recently,~\cite{vp2} conducted a simulated benchmark for action-conditioned video prediction in the form of an MPC framework that evaluated a given model for simulated robotic manipulation through sampling-based planning.

Recently, some video prediction models independent of the MPC framework have also been proposed for robotic manipulation. For instance, VLP~\citep{du2024video} and UniPi~\citep{du2023learning} combined text-to-video models with VLM to generate long-horizon videos used for extracting control actions. V-JEPA ~\citep{bardes2024vjepa} developed a latent video prediction strategy to make predictions in a learned latent space. Similarly, Dreamer~\citep{Hafner2020Dream} learned long-horizon behaviors by predicting state values and actions in a compact latent space where the latent states have a small memory footprint. RIG~\citep{nair2018visual} used a latent variable model to generate goals for the robot to learn diverse behaviors. Planning~to~Practice~\citep{fang2022planning} proposed a sub-goal generator to decompose a goal-reaching task hierarchically in the latent space.

\subsection{Foundation Models for Robotic Manipulation}
Foundation models are large-scale neural networks trained on massive and diverse datasets~\citep{bommasani2021opportunities}. Breakthroughs such as GPT-4, Llama and PaLM exemplify the scaling up of LLMs~\citep{gpt4,gpt3,llama,palm}, showcasing notable progress in knowledge extraction and reasoning. Simultaneously, there has been an increase in the development of large-scale VLMs~\citep{flamingo,clip,align,dalle,palm-e,qwen}. VLMs typically employ cross-modal connectors to merge visual and textual embeddings into a unified representation space, enabling them to process multimodal data effectively. 

The application of foundation models in advanced robotic systems is an emerging research field. Many studies focus on the use of LLMs for knowledge reasoning and robotic manipulation~\citep{huang2022inner,zeng2022socratic,huang2023grounded,liang2023code,hu2023look}. To allow LLMs to perceive physical environments, auxiliary modules such as textual descriptions of the scene~\citep{huang2022inner,zeng2022socratic}, affordance models~\citep{huang2023grounded}, and perception APIs~\citep{liang2023code} are essential. Furthermore, the use of VLMs for robotic manipulation has been explored~\citep{voxposer,palm-e,rt-2}. For example, PaLM-E enhanced the understanding of robots with regard to complex visual-textual tasks~\citep{palm-e}, while RT-2 specialized in real-time image processing and decision making~\citep{rt-2}. However, most existing methods are limited by their reliance on pre-defined executable skills or hand-designed motion primitives~\citep{liang2023code,voxposer}, constraining the adaptability of robots in complex, real-world environments and their interaction with diverse, unseen objects.

\textbf{Difference from closely related work.} This work is closely related to some MPC-based methods~\citep{visual_foresight,deep_visual_foresight,vp2,3dmpc} designed for robotic manipulation. However, most of these methods were designed for manipulation tasks merely involving specific objects as regular MPC has limitations in two aspects: (1) The predictive models used in regular MPC are constrained with small-scale training datasets, and thus cannot precisely predict the process of interaction with objects unseen during training; (2) The cost functions of regular MPC are usually designed with a defined set of constraints such as physical limitations or operational safety margins. Although these constraints ensure that robot actions adhere to them while striving for optimal performance, accurately modeling such constraints is highly difficult in real-world scenarios. To address the above two problems, the proposed VLMPC leverages a video prediction model trained with a large-scale robot manipulation dataset~\citep{padalkar2023open} and can be directly transferred to the real world. Also, VLMPC incorporates powerful VLMs into cost functions with high-level knowledge reasoning, which provides constraints produced through interactions with open-set objects.


Different from directly predicting executable actions, another approach to integrating foundation models in robotic manipulation is to predict motion trajectories. \cite{xu2024flowcrossdomainmanipulationinterface} introduced a flow-generation model that encodes language instructions using CLIP \citep{clip} to generate object flow as a robotic manipulation interface, followed by a flow-conditioned policy to determine robot actions. \cite{yuan2024general} proposed a language-conditioned 3D flow prediction model trained on large-scale RGB-D human video datasets, leveraging object flow predictions in 3D scenes for manipulation tasks. In contrast to object flow approaches, \cite{wen2023anypoint} pre-trained a trajectory model using video demonstrations to predict future trajectories of any point in a frame, facilitating the manipulation of articulated and deformable objects.
However, these methods face challenges in accurately transforming 2D flows into executable 3D trajectories, particularly concerning the robot's z-axis movements in camera coordinates. Additionally, their performance relies heavily on the precision of flow predictions, which depends strongly on the scale and diversity of training data, limiting their ability to generalize to unseen objects and environments. To address such limitations, the proposed Traj-VLMPC introduces a trajectory-based approach that leverages a Gaussian Mixture Model (GMM) for adaptive 3D trajectory sampling. Unlike prior works that rely solely on flow predictions, Traj-VLMPC integrates VLM-based spatial reasoning with probabilistic motion modeling, ensuring more reliable trajectory generation even in novel scenarios. Furthermore, by incorporating a voxel-based 3D value map for trajectory assessment, Traj-VLMPC improves planning efficiency and collision awareness, offering a more robust solution for complex manipulation tasks.

\section{Method}
\label{sec:methods}
Fig.~\ref{fig1} illustrates the overview of the VLMPC framework. It takes as input either a goal image indicating the prospective state or a language instruction that depicts the required manipulation, and performs a dynamic strategy that iteratively makes decisions based on the predictions of future frames. First, a conditional action sampling scheme is designed to prompt VLMs to take into account both the input and the current observation and reason out prospective future movements, from which a set of candidate action sequences are sampled. Then, an action-conditioned video prediction model is devised to predict a set of future frames corresponding to the sampled action sequences. Finally, a hierarchical cost function including two sub-costs and a VLM switcher are proposed to comprehensively compute the coarse-to-fine scores for the video predictions and select the best action sequence. The first action in the sequence is fed into the robot for execution, and the subsequent actions go through a weighted elementwise summation with the conditional action distribution. We elaborate each component of VLMPC in the following.

\subsection{Conditional Action Sampling}
In an MPC framework, $N$ candidate action sequences $\mathcal{S}_t = \{S^1_t, S^2_t, ..., S^N_t\}$ are sampled from a custom sampling distribution at each step $t$, where $S^n_t = \{a^{n}_{t+1}, a^{n}_{t+2}, ..., a^{n}_{t+T}\}$ contains $T$ actions and $n \in \{1,...,N\}$. For every $\tau \in \{t+1,...,t+T\}$ representing a future step after $t$, $a^{n}_{\tau} \in \mathbb{R}^7$ is a $7$-dimensional vector composed of the movement $[d^x_\tau, d^y_\tau, d^z_\tau]$ of the end-effector in Cartesian space, the rotation $[r^x_\tau, r^y_\tau, r^z_\tau]$ of the gripper, and a binary grasping state $g_t$ indicating the open or close state of the end-effector. 

\begin{figure}[t]
    \centering
    \includegraphics[width=\linewidth]{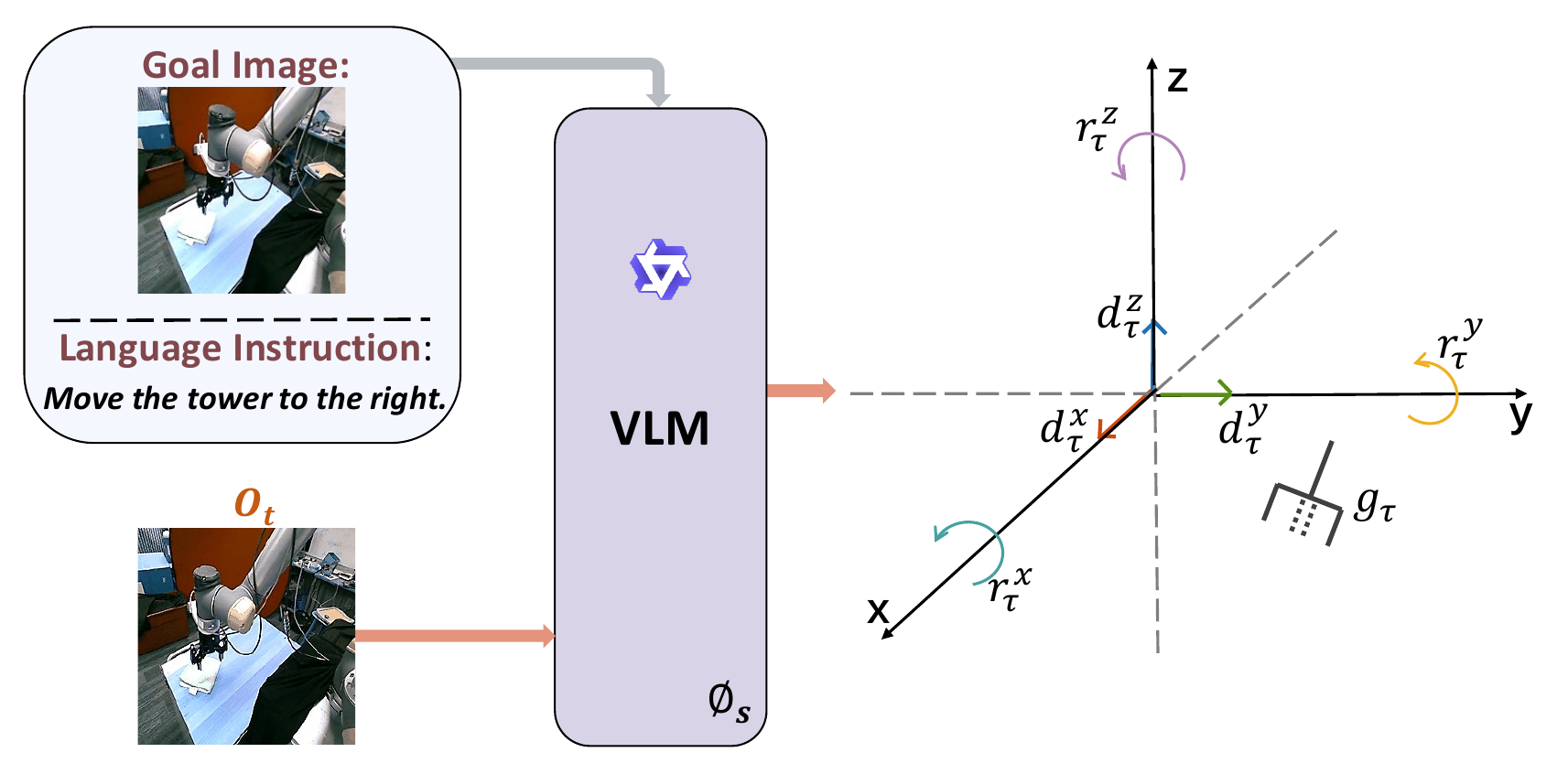}
    \caption{The VLMs subject to a specifically designed prompt $\phi_s$ take as input the current observation $O_t$ and a goal image or a language instruction to generate an end-effector moving direction at coarse level.} 
    \label{fig2}
\end{figure}

Given a goal image $G$ or a language instruction $L$ as the input of VLMPC along with the current observation $O_t$, we expect VLMs to generate appropriate future movements, from which a sampling distribution is derived for action sampling. As shown in Fig.~\ref{fig2}, the current observation $O_t \in \mathbb{R}^{h\times w \times 3}$ is represented as an RGB image with the shape of $h\times w \times 3$ taken by an external monocular camera. We design a prompt $\phi_s$ that drives VLMs to analyze $O_t$ alongside the input. $\phi_s$ forces VLMs to identify and localize the object with which the robot is to interact, reason about the manner of interaction, and generate appropriate future movements. The output of VLMs can be formulated as
\begin{equation}
\label{eq1}
    \text{VLM}(O_t, G  \lor L|\phi_s) = \{\widehat{d}^x_t, \widehat{d}^y_t, \widehat{d^z_t}, \widehat{r^x_t}, \widehat{r^y_t}, \widehat{r^z_t}, g_t\} 
\end{equation}
where $\widehat{\cdot} \in \{+1, 0, -1\}$ denotes the predicted moving/rotation direction alongside the corresponding axis and $g_t \in \{0, 1\}$ represents the predicted binary state of the end-effector.

To obtain a set of candidate action sequences, we follow the scheme of Visual Foresight~\citep{visual_foresight} and adopt Gaussian sampling that samples $N$ action sequences with the expected movement in each action dimension as the mean. Hence we further map the output of VLMs into a sampling mean $\mu^\text{VLM}_t$:
\begin{equation}
\label{eq2}
    \mu^\text{VLM}_t = w_m * \{\widehat{d}^x_t, \widehat{d}^y_t, \widehat{d}^z_t\} \cup w_r * \{\widehat{r}^x_t, \widehat{r}^y_t, \widehat{r}^z_t\} \cup \{g_t\}
\end{equation}
where $w_m$ and $w_r$ are the hyperparameters for mapping the output of VLMs into the action space of the robot.

Hallucination phenomenon is a common issue which hinders the stable use of large-scale VLMs in real-world deployment, as it may
result in unexpected consequences caused by incorrect understandings of the external environment. To mitigate the hallucination phenomenon, we propose to make use of the historical information derived from the subsequent candidate action sequence of the last step. This leads to another sampling mean $\mu^\text{sub}_t$. Please refer to Sec.~\ref{cost} for the detailed process of obtaining $\mu^\text{sub}_t$. Then we perform a weighted elementwise summation of $\mu^\text{sub}_t$ and $\mu^\text{VLM}_t$ to produce the final sampling mean $\mu_t$ of step $t$:
\begin{equation}
\label{eq3}
    \mu_t =w_\text{VLM} * \mu^\text{VLM}_t + w_\text{sub} * \mu^\text{sub}_t
\end{equation}
where $w_\text{VLM}$ and $w_\text{sub}$ are weighting parameters. Finally, we sample $S_t$ from the Gaussian distribution $S_t^n\sim\mathcal{N}(\mu_t, I)$ repeatedly $N$ times.

This conditional action sampling scheme allows VLMs to provide the guidance of robotic manipulation at a coarse level via knowledge reasoning from the image observation and the task goal. Next, with the candidate action sequences, we introduce the module for action-conditioned video prediction.

\subsection{Action-Conditioned Video Prediction}

Given the candidate action sequences, it is necessary to estimate the future state of the system when executing each sequence, which provides the forward-looking capability of VLMPC.

Traditional MPC methods often rely on hand-crafted deterministic dynamic models. Developing and refining such models typically requires extensive domain knowledge, and they may not capture all relevant dynamics. On the contrary, video is rich in semantic information and thus enables the model to handle complex, dynamic environments more effectively and flexibly. Moreover, video can be directly fed into a VLM for knowledge reasoning. Thus, we use the action-conditioned video prediction model to predict the future frames corresponding to candidate action sequences. 

\begin{figure}[t]
    \centering
\includegraphics[width=\linewidth]{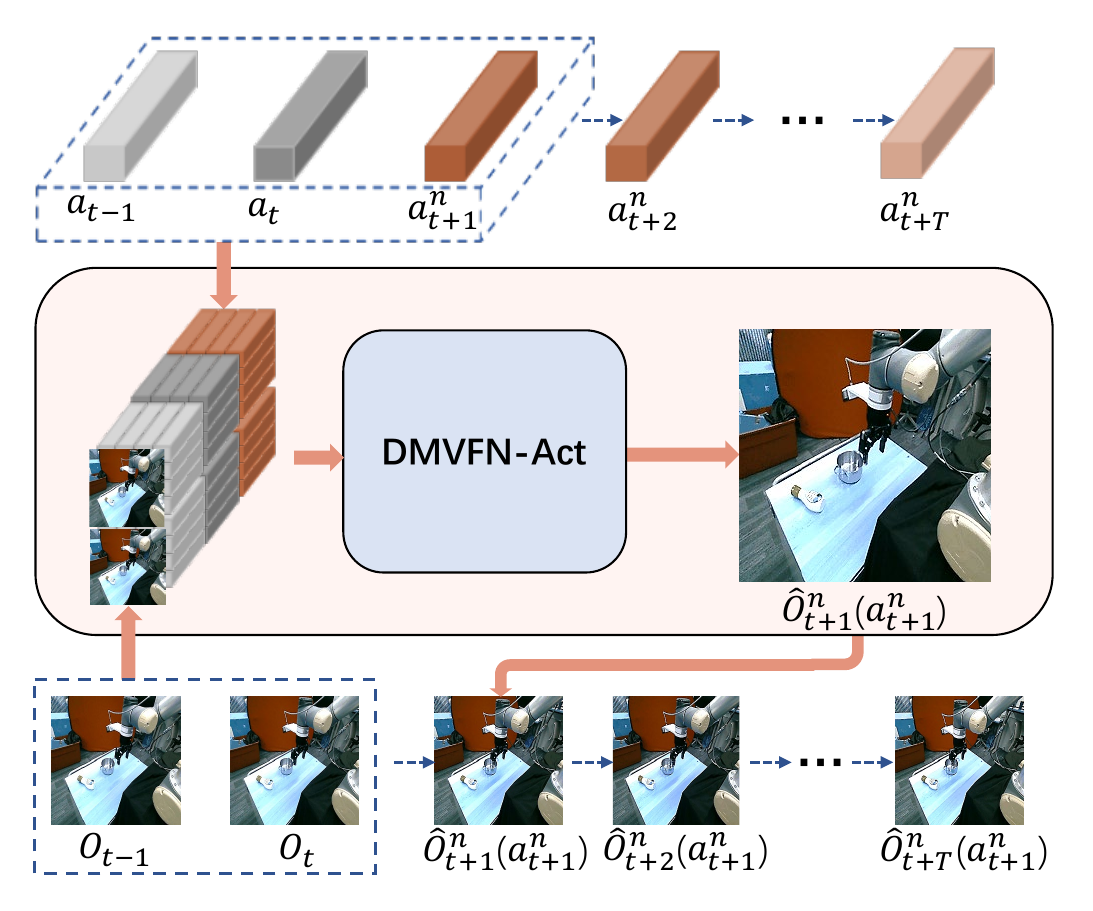}
    \caption{Given the past two frames $O_{t}$ and $O_{t-1}$ with the executed actions $a_{t-1}$ and $a_{t}$ corresponding to them and the action $a^{n}_{t+1}$,  DMVFN-Act predicts the next frame $\widehat{O}^{n}_{t+1}(a^{n}_{t+1})$. The dashed boxes and arrows indicate the iterative process of taking the actions one by one and predicting the future states frame by frame.}
    \label{fig3}
\end{figure}

We build a variant version of DMVFN~\citep{dmvfn}, an efficient dynamic multi-scale voxel flow network for video prediction, to perform action-conditioned video prediction. We name it DMVFN-Act. Given the past two historical frames $O_{t-1}$ and $O_t$, DMVFN predicts a future frame $\widehat{O}_{t+1}$, formulated as
\begin{equation}
\label{eq4}
    \widehat{O}_{t+1} = \text{DMVFN}(O_{t-1}, O_{t}).
\end{equation}

With the candidate action sequences $S_t$ and the corresponding executed actions $a_{t-1}$ and $a_t$, we expect DMVFN-Act to take the actions one by one and predict future states frame by frame as illustrated in Fig.~\ref{fig3}. For simplicity, we explain this process by taking one sequence $S^n_t = \{a^{n}_{t+1}, a^{n}_{t+2}, ..., a^{n}_{t+T}\}$ as example. We broadcast $a_{t-1}$, $a_t$, $a^{n}_{t+1} \in \mathbb{R}^7$ to the image size $a_{t-1}'$, $a_t'$, ${a^{n}_{t+1}}' \in \mathbb{R}^{h\times w\times 7}$, and then concatenate them 
with  $O_{t-1}$ and $O_t$ respectively, formulated as
\begin{equation}
\begin{split}
    O_{t-1}' &= [O_{t-1}\cdot a_{t-1}' \cdot a_t' \cdot {a^{n}_{t+1}}'], \\
    O_{t}' &= [O_{t}\cdot a_{t-1}' \cdot a_t' \cdot {a^{n}_{t+1}}']
     \label{eq5}
\end{split}
\end{equation}
where $[\cdot]$ represents the channelwise concatenation, and $O_{t-1}'$ and $O_{t}'$ denote the action-conditioned historical observations. In DMVFN-Act, the input layer is modified to adapt $O_{t-1}'$ and $O_{t}'$ and predict one future frame $\widehat{O}^{n}_{t+1}(a^{n}_{t+1})$ conditioned by the candidate action $a^{n}_{t+1}$, expressed as
\begin{equation}
    \widehat{O}^{n}_{t+1}(a^{n}_{t+1}) = \text{DMVFN-Act}(O_{t-1}', O_{t}').
    \label{eq6}
\end{equation}
DMVFN-Act iteratively predicts future frames via Eqs.~(\ref{eq5}) and  (\ref{eq6}) until all candidate actions are used. The action-conditioned video prediction can be represented as:
\begin{equation}
\label{eq7}
    V^n_t = \{
    \widehat{O}^{n}_{t+1}(a^{n}_{t+1}), \widehat{O}^{n}_{t+2}(a^{n}_{t+2}),..., \widehat{O}^{n}_{t+T}(a^{n}_{t+T})\}.
\end{equation}
To improve efficiency, the $N$ candidate action sequences are organized into a batch and predict all the action-conditioned videos $V_t=\{V^1_t, V^2_t, ..., V^N_t\}$ at step $t$ in one inference.

\subsection{Hierarchical Cost Function}\label{cost}
To comprehensively assess the video predictions, we design a cost function composed of two sub-costs that provide a hierarchical assessment at pixel and knowledge levels, respectively. We also propose a VLM switcher which dynamically selects one or both sub-costs in a manner adaptive to the observation.

\subsubsection{Pixel Distance Cost.}
While the task input is the goal image $G$, an intuitive way to assess video predictions is to sum the pixel distances between each future frame and the goal image. Following Visual Foresight~\citep{visual_foresight}, we calculate the $l_2$ distance between each future frame $\widehat{O}^{n}_{\tau}(a^{n}_{\tau})$ in an action-conditioned video $V_t^n$ and $G$, and then sum the distances as the pixel distance cost $C_P^n(t)$ for $V_t^n$ over $\tau$:
\begin{equation}
\label{eq8}
    C_P^n(t)=\sum\limits_{\tau=t+1}^{t+T}||\widehat{O}^{n}_{\tau}(a^{n}_{\tau})-G||_2.
\end{equation}
Then, the pixel distance cost $C_P(t)$ at step $t$ for $V_t$ can be computed as
\begin{equation}
\label{eq9}
    C_P(t)=\{C_P^n(t) |n\in \{1,2,..., N\}\}.
\end{equation}

\begin{figure}[t]
    \centering
    \includegraphics[width=\linewidth]{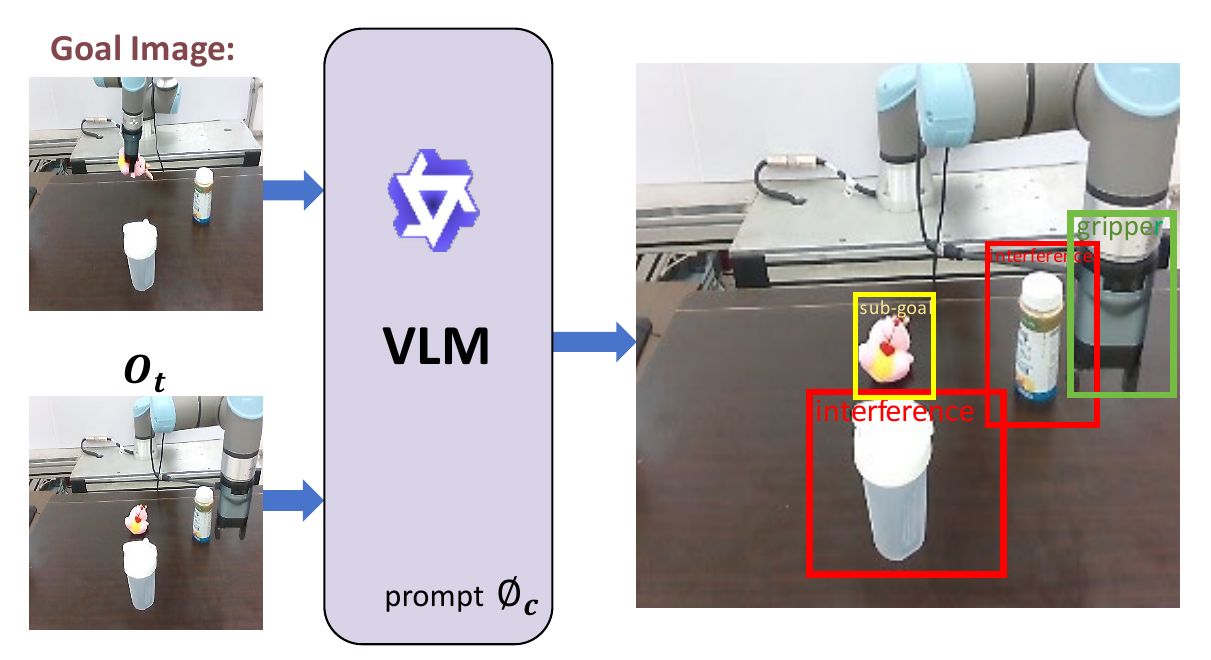}
    \caption{Illustration of the end-effector, the next sub-goal and the
interference objects in the current observation. Red, green, and yellow boxes denote the interference objects, the end-effector and the next sub-goal generated by VLMPC.}
    \label{fig4}
\end{figure}

The pixel distance cost encourages the robot to move directly towards the goal position in accordance with the goal image. This cost is simple yet effective when the task contains only one sub-goal, \eg \textit{push a button}. However, for tasks that require manipulating objects with multiple sub-goals, where a common type is \textit{taking an object from position A to B}, this cost usually guides the robot to move directly towards \textit{position B} to reduce the pixel distance. To facilitate such situations, we further introduce the VLM-assisted cost. 

\subsubsection{VLM-Assisted Cost.}
Many robotic manipulation tasks contain multiple sub-goals and interference objects, which require knowledge-level task planning. For example, in the task of \textit{grasp the bottle and put it in the bowl, while watching out the cup}, the bottle should be identified as the sub-goal before the robot grasps it, and the bowl is the next sub-goal after the bottle is grasped, where the cup is an interference object. 
It is thus critical to dynamically identify the sub-goals and interference objects in each step, and make appropriate assessments on the action-conditioned video predictions so that we can select the best candidate action sequence to achieve the sub-goals while avoiding the interference object. We design a VLM-assisted cost to realize it at the knowledge level.

\begin{algorithm}[t]
    \SetAlgoNoLine
    \caption{VLMPC}
    \label{algo}
    \KwIn{Goal image $G$ or language instruction $L$, and obvevation $O_t$ at every step}
    \BlankLine
    \While{task not done $\text{\textbf{or}}$ $t\leq T_{max}$}{
        Generates a sampling distribution by VLM
        $D(\mu^{\text{VLM}}) \leftarrow \text{VLM}(O_t, G  \lor L|\phi_s)$\;
        Refine it with historical information $\mu^\text{sub}_t$ \
        $D(\mu_t) = D(w_\text{VLM} * \mu^\text{VLM}_t + w_\text{sub} * \mu^\text{sub}_t$)\;
        $\mathcal{S}_t \leftarrow$ sample $N$ action sequences\;
        \ForEach{sequence $S^n_t \in \mathcal{S}_t$}{
            \For{future step $\tau=t+1,...,t+T$}{
                $\widehat{O}^{n}_{\tau}(a^{n}_{\tau}) \leftarrow$ predict the future frame\;
            }
            $V^n_t=\{\widehat{O}^{n}_{\tau}(a^{n}_{\tau})|\tau \in \{t+1,...,t+T\}$ \;
        }
        $C_P(t) \leftarrow$ calculate the pixel distance cost\;
        $C_\text{VLM} \leftarrow$ calculate the VLM-assisted cost\;
        $C_t{ \leftarrow}$ arrange cost through VLM swicher\;
        $S^{n^{\star}}_t \leftarrow$ select the optimal action sequence\;
        Execute the first action $a^{n^{\star}}_{t+1}$ in the optimal sequence\;
        Update $\mu_{t+1}^{sub}$ using $\{a^{n^{\star}}_{\tau}|\tau \in \{t+2,...,t+T\}\}$\;
    }
\end{algorithm}

Specifically, with the current observation $O_t$ and the task input $G$ or $L$, we design a prompt $\phi_C$ to drive VLMs to reason out and localize the next sub-goal and all the interference objects, where the sub-goal is usually the next object to interact with the robot. As shown in Fig.~\ref{fig4}, this process yields the bounding boxes of the robot's end-effector $e_t$, the next sub-goal $s_t$, and all the interference objects $I_t$ in the current observation:
\begin{equation}
\label{eq10}
    \text{VLM}(O_t, G \lor L | \phi_C) = \{e_t, s_t, I_t\}.
\end{equation}
Since the predicted videos $V_t$ share the historical frame $O_t$, a lightweight visual tracker VT can be used to localize both the end-effector $e_\tau^n$ and the sub-goal $s_\tau^n$ in each future frame in all the action-conditioned videos initialized with $e_t$, $s_t$, and $I_t$, formulated as:
\begin{equation}
\label{eq11}
\begin{split}
    \text{VT}(V_t | e_t, s_t, I_t) = \{&e_\tau^n, s_\tau^n, I_\tau^n | n\in \{1,2,...,N\}, \\ &\tau \in \{t+1, t+2, ..., t+T\}\}
\end{split}
\end{equation}
where we employ an efficient real-time tracking network SiamRPN~\citep{siamrpn} as the visual tracker in this work.

To encourage the robot to move towards the next sub-goal and avoid colliding with all the interference objects, we calculate the VLM-assisted cost $C_\text{VLM}^n$ as:
\begin{equation}
\label{eq12}
\begin{split}
    C_\text{VLM}^n(t) = \sum\limits_{\tau=t+1}^{t+T}(&||c(e_\tau^n) - c(s_\tau^n)||_2 \\-&||c(e_\tau^n)-c(I_\tau^n)||_2),
\end{split}
\end{equation}
\begin{equation}
\label{eq13}
C_\text{VLM}(t)=\{C_\text{VLM}^n(t)|n\in\{1,2,...,N\}\}
\end{equation}
where $c(\cdot)$ represents the center of the bounding box.


 

\subsubsection{VLM Switcher.}
The pixel distance cost can provide fine-grained guidance on the pixel level, and the VLM-assisted cost fixes the gap in knowledge-level task planning. Based on the two sub-costs, we further design a VLM switcher with prompt $\phi_{D}$, which dynamically selects one or both appropriate sub-costs in each step $t$ adaptive to the current observation through knowledge reasoning to produce the final cost $C(t)$: 
\begin{equation}
\label{eq14}
    \text{VLM}(O_t, G\lor L|\phi_D) = w_D \in \{0,0.5,1\},
\end{equation}
\begin{equation}
\label{eq15}
    C(t) = w_D * C_P(t) + (1-w_D) * C_\text{VLM}(t).
\end{equation}
With the cost $C(t) = \{C^n(t)|n\in\{1,2,...,N\}\}$ as the assessment of all the action-conditioned videos, we select the candidate action sequence with the lowest cost for the following process. When the first action in this sequence is executed, the subsequent actions are fed into a global mean pooling layer to generate the sampling mean $\mu_t^{sub}$ to provide historical information in the action sampling of the next step.

Algorithm~\ref{algo} summarizes the whole process of the VLMPC framework. When the task is done or reaching the maximum time limit, the system will return an end signal.

\section{Traj-VLMPC: A Trajectory-Based Variant of VLMPC}
\label{sec:traj}

\begin{figure}[t]
    \centering
    \includegraphics[width=1\linewidth]{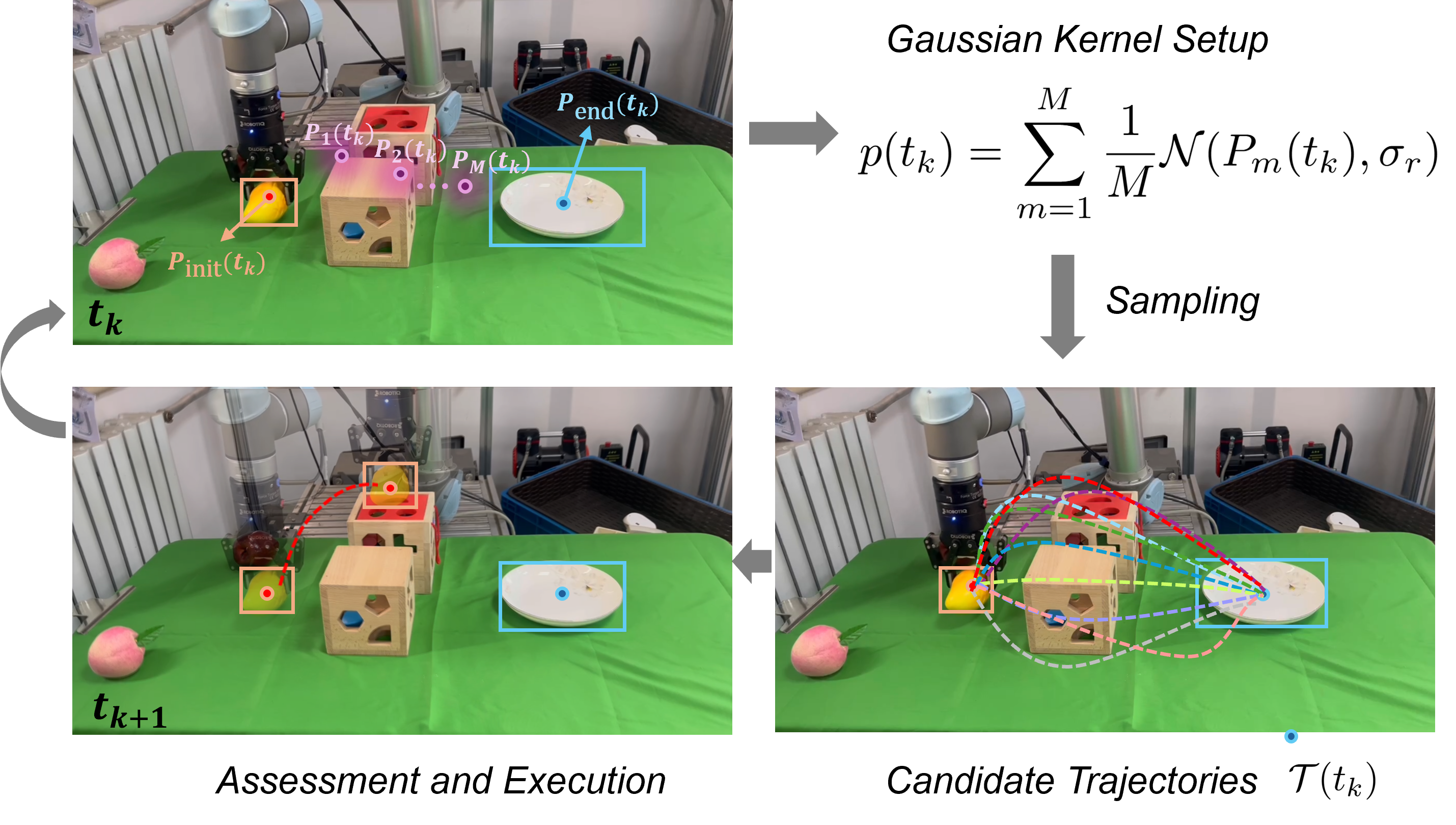}
    \caption{Workflow of Traj-VLMPC. Given the end-effector position $P_\text{init}(t_k)$ and the sub-goal $P_\text{end}(t_k)$, a GMM $p(t_k)$ is constructed in 3D space with $M$ kernels. Candidate trajectories are sampled from the GMM and evaluated via the voxel-based 3D value map, with the lowest-cost path executed at each time step in an MPC loop.}
    \label{fig_vlmpc}
\end{figure}

Although VLMPC showcases the potential to unify VLM and MPC into a cohesive framework, its reliance on step-by-step video prediction and VLM inference results in substantial computational costs. To overcome this limitation while preserving action accuracy, we introduce Traj-VLMPC, an enhanced variant of VLMPC that shifts the focus from single-action sampling and evaluation to motion trajectories. As shown in Fig.~\ref{fig_vlmpc}, we introduce a Gaussian Mixture Model (GMM)-based trajectory sampling strategy, which generates a set of 3D proposal trajectories for the end-effector guided by VLM. Additionally, we extend the assessment process from 2D frames to a 3D affordance map, enabling a more efficient and comprehensive evaluation.


\subsection{Trajectory Sampling with GMM
}
\label{sec:4.1}

The computational burden of step-by-step action sampling following video prediction and evaluation in VLMPC limits its efficiency significantly. Motivated by recent advances in probabilistic motion modeling and trajectory optimization~\citep{voxposer,im2flow2act,orion}, we introduce Traj-VLMPC, which employs a GMM-based trajectory sampling strategy to generate diverse and feasible motion trajectories for the end-effector in 3D space.  

Unlike the step-by-step action sampling in VLMPC, Traj-VLMPC employs a trajectory-based sampling strategy, where trajectory sampling occurs only at discrete time steps, given by $t_k=t\cdot (1/f)$, with a pre-defined frequency $f$. Given the image observation and the language instruction, the prompted VLM first localizes the end-effector $e_{t_k}$, the next sub-goal $s_{t_k}$, and all the interference objects $I_{t_k}$ in the image at time step $t_k$ similar to VLMPC in Eq.~(\ref{eq10}). 
The setup of Gaussian kernels and the process of trajectory sampling are introduced in detail below. 

\subsubsection{Gaussian Kernel Setup.}
To ensure an adaptive motion trajectory generation, we construct Gaussian kernels in 3D space which serve as probabilistic representations of spatial uncertainty and guide the sampling of feasible trajectory candidates.

We define the initial trajectory point $P_\text{init}(t_k)$ and the ending trajectory point $P_\text{end}(t_k)$ at $t_k$ as the centers of the end-effector and the sub-goal in 3D space respectively, expressed as:
\begin{equation}
    P_\text{init}(t_k) = \mathbf{T}c(e_{t_k})    
\end{equation}
and
\begin{equation}
    P_\text{end}(t_k) = \mathbf{T}c(s_{t_k})
\end{equation}
where \(\mathbf{T}\) is the transformation matrix that maps 2D image coordinates to 3D space, derived from RGB-D sensor calibration and camera intrinsic parameters. Subsequently, we randomly sample $M$ intermediate points between $P_\text{init}(t_k)$ and $P_\text{end}(t_k)$, which serve as the centers of the Gaussian kernels in 3D space. These sampled points are given by:
\begin{equation}
\begin{split}
    P_m(t_k) = P_\text{init}(t_k) + \lambda_m (P_\text{end}(t_k) - P_\text{init}(t_k))
\end{split}
\end{equation}
where $m\in\{1,2,...,M\}$. $\lambda_m$ follows a uniform distribution $\mathcal{U}(0,1)$, ensuring that the sampled points are evenly distributed along the trajectory segment. These points serve as the mean positions of the 3D Gaussian kernels, capturing the spatial uncertainty in trajectory sampling.  

Each sampled trajectory point \( P_m(t_k) \) is modeled as a 3D Gaussian kernel, and jointly composes the following 3D Gaussian distribution as:
\begin{equation}
    p(t_k)=\sum_{m=1}^{M}\frac{1}{M}\mathcal{N}(P_m(t_k),\sigma_r)
\end{equation}
where the hyperparameter $\sigma_r$ controls the variance of the initial distribution, regulating the uncertainty in the end-effector’s starting position within the proposal trajectories.

\subsubsection{Trajectory Sampling.}
With the 3D Gaussian distribution, we aim to produce $J$ candidate trajectories. For the $j$-th candidate trajectory at step $t_k$, we first iteratively sample a subset of $N_\text{sub}$ trajectory points:
\begin{equation}
    \mathcal{T}^j_{\text{sub}}(t_k) = \{P_i(t_k)\sim p(t_k)|i\in\{1,2,...,N_\text{sub}\} \}.
\end{equation}
Then, we employ linear interpolation to generate a smooth and continuous candidate trajectory by interpolating additional points between the sampled trajectory subset:
\begin{equation}
    \mathcal{T}^j(t_k) = \text{Interpolate}(\mathcal{T}^j_{\text{sub}}(t_k))
\end{equation}
where Interpolate represents a linear interpolation function that refines the trajectory by computing intermediate points between adjacent sampled points. This process ensures that the generated candidate trajectory is smooth and evenly distributed, facilitating efficient motion planning while preserving the underlying probabilistic structure of the 3D Gaussian sampling.

At each time step \( t_k \), we repeat the above trajectory sampling process \( J \) times, resulting in a set of \( J \) candidate trajectories:
\begin{equation}
    \mathcal{T}(t_k) = \{\mathcal{T}^j(t_k) | j \in \{1,2,...,J\} \}.
\end{equation}
These candidate trajectories serve as motion hypotheses and will be further evaluated within the MPC loop to select the optimal control sequence. The candidate trajectories are then passed into the MPC loop, where they undergo further evaluation to determine the optimal control sequence, ensuring efficient and goal-directed robotic execution.

\subsection{Trajectory Assessment and Execution}
\label{sec:4.2}
Consistent with VLMPC, the MPC loop of Traj-VLMPC requires further evaluation of candidate trajectories $\mathcal{T}(t_k)$ to select the optimal trajectory for execution. 
Recently, VoxPoser~\citep{voxposer} constructs a voxel-based 3D value map to provide a structured and spatially-aware representation of task constraints. This 3D value map effectively encodes task-relevant affordances and obstacles, where high-value regions guide the end-effector to move toward target objects, and low-value regions indicate areas that need to be avoided. Inspired by such a spatially grounded method, we integrate a voxel-based 3D value map into the MPC loop to assess candidate trajectories in Traj-VLMPC.

Different from VoxPoser, which employs LLMs to compose code by querying a VLM and constructing a voxel-based 3D value map, Traj-VLMPC directly uses the spatial information extracted from the VLM in previous sampling stages to eliminate the need for redundant queries. Given the position of the sub-goal $c(s_{t_k})$ and all interference objects $c(I_{t_k})$, we aim to construct a voxel-based 3D value map $\mathbf{V}\in\mathbb{R}^{w\times h\times d}$ that indicates regions favorable for trajectory execution while penalizing areas with potential collisions or task constraints.

Specifically, we first initialize a 3D voxel-based value map $\mathbf{V}$ within the operational space of the robotic arm, where each voxel $\mathbf{x}$ is initially set to zero as $\mathbf{V}(\mathbf{x})=0$. Then we assign a value of $-1$ to the voxel corresponding to the sub-goal position:
\begin{equation}
    \mathbf{V}(\mathbf{T}c(s_{t_k})) = -1
\end{equation}
and a value of $1$ to the voxels corresponding to the positions of interference objects:
\begin{equation}
    \mathbf{V}(\mathbf{T}c(I_{t_k})) = 1.
\end{equation}

\begin{figure}[t]
    \centering
    \includegraphics[width=1\linewidth]{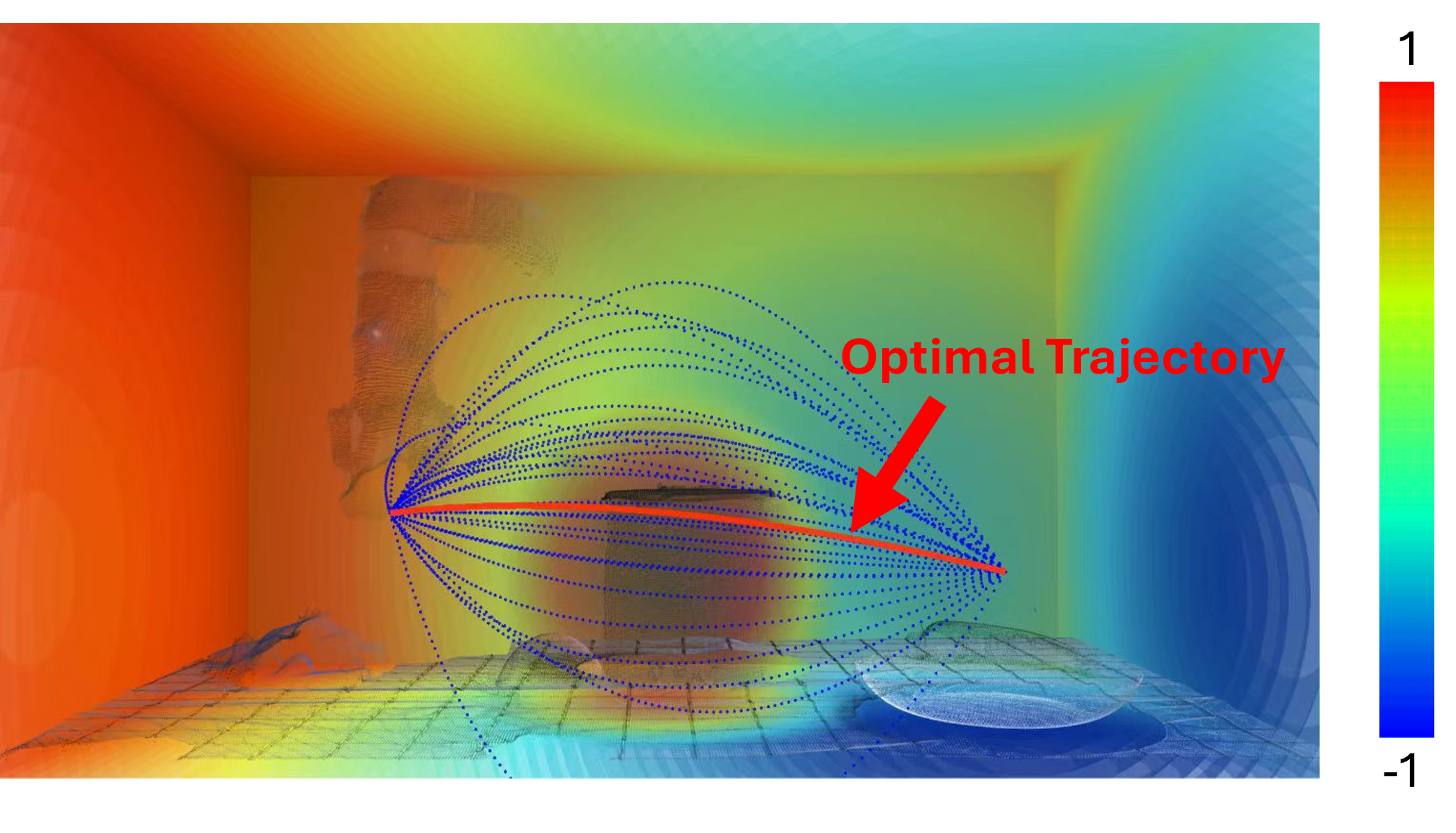}
    \caption{Visualization of the GMM-based 3D value map. The color scale from $-1$ (blue) to $1$ (red) represents cost values, where lower scores favor sub-goal regions and higher scores indicate interferences. By summing these values along each candidate trajectory, Traj-VLMPC selects the path with the lowest cost as the optimal option for robotic execution.}
    \label{fig_valuemap}
\end{figure}

To ensure a smooth spatial representation of task constraints, we apply Gaussian spreading to propagate values across the 3D voxel space. The value at each voxel \( \mathbf{x} \) is updated using a Gaussian-weighted distance function:
\begin{equation}
\begin{split}
    \mathbf{V}(\mathbf{x}) = -&\exp\left(-\frac{\|\mathbf{x} - \mathbf{T}c(s_{t_k})\|^2}{2\sigma_s^2}\right) 
    + \\
    &\sum_{j} \exp\left(-\frac{\|\mathbf{x} - \mathbf{T}c(I_{t_k}^j)\|^2}{2\sigma_I^2}\right)
\end{split}
\end{equation}
where $\sigma_s$ and $\sigma_I$ control the spread of the sub-goal and the interference object influence, respectively. As shown in Fig.~\ref{fig_valuemap}, this Gaussian spreading process ensures a smooth transition between high-risk areas (near obstacles) and goal-attractive regions.

Based on the 3D voxel value map, we assess each candidate trajectory by accumulating the value map scores along its trajectory points. For a given candidate trajectory $\mathcal{T}^j(t_k)$ consisting of $N_\mathcal{T}$ discrete points $P_i^j(t_k)$, the trajectory cost is computed as:
\begin{equation}
    C_j = \sum_{i=1}^{N_\mathcal{T}} \mathbf{V}(P_i^j(t_k)).
\end{equation}
A lower cost indicates a trajectory that better aligns with the task objective by favoring goal-reaching regions and avoiding interference objects. We evaluate all candidate trajectories and select the trajectory with the lowest cost as the optimal trajectory for execution:
\begin{equation}
    \mathcal{T}^* (t_k) = \text{argmin}_{\mathcal{T}^j(t_k) \in \mathcal{T}(t_k)} C_j.
\end{equation}
This cost-driven trajectory assessment ensures that the executed trajectory is both task-efficient and collision-aware, leveraging the structured spatial information encoded in the 3D voxel value map.

Following VoxPoser~\citep{voxposer}, we adopt a closed-loop trajectory execution strategy, where the robot iteratively refines its motion based on real-time perception feedback. At each time step \( t_k \), the selected optimal trajectory \( \mathcal{T}^*(t_k) \) is executed incrementally, allowing the system to dynamically adjust to correct potential deviations. In the MPC loop, we continuously update the 3D voxel value map at each step $t_k$. This allows the MPC loop to re-evaluate and select the optimal candidate trajectory, ensuring that the motion remains optimal despite variations in the scene.




\section{Experiments}
\label{sec:experiments}
In this section, we first provide the implementation details of the proposed VLMPC framework. Then, we conduct two comparative experiments in simulated environments. The first is to compare VLMPC with VP$^2$ \citep{vp2} on 7 tasks in the RoboDesk environment \citep{kannan2021robodesk}. The second is to compare VLMPC with 5 existing methods in 50 simulated environments provided by the Language Table environment \citep{lynch2023interactive}. Then, we evaluate VLMPC and Traj-VLMPC in real-world scenarios. Finally, we investigate the effectiveness of each core component of VLMPC through ablation studies. In the supplementary material, we provide the details of all the hyperparameters and the VLM prompts.


\subsection{Implementation Details}
VLMPC employs Qwen-VL~\citep{qwen} and GPT-4V~\citep{2023GPT4VisionSC} as VLMs. In the conditional action sampling module, VLMPC first uses GPT-4V to identify the target object with which the robot needs to interact, and then localizes the object through Qwen-VL. In the VLM-assisted cost, VLMPC first extracts sub-goals and interference objects with GPT-4V, and then localizes them through Qwen-VL. The VLM switcher uses GPT-4V to dynamically select one or both sub-costs in each time step. In Traj-VLMPC, we use GPT-4V and DINO-X~\citep{ren2024dinoxunifiedvisionmodel} as VLMs. Similar to VLMPC, Traj-VLMPC initially uses GPT-4V to identify the sub-goal, followed by DINO-X for precise sub-goal localization. In the process of voxel-based trajectory assessment, Traj-VLMPC first recognizes the sub-goal and all the potential interference objects, and then localizes them through DINO-X. 

The training policy of the DMVFN-Act video prediction model contains 2 stages. In the first stage, we select 3 sub-datasets from the Open X-Embodiment Dataset~\citep{padalkar2023open}, a large-scale dataset containing more than 1 million robot trajectories collected from 22 robot embodiments. The 3 sub-datasets used for pre-training DMVFN-Act are Berkeley Autolab UR5, Columbia PushT Dataset, and ASU TableTop Manipulation. In the second stage, we collect 20 episodes of robot execution in the environment where the experiments are conducted and train DMVFN-Act to adapt to the specific scenario.

\begin{figure*}[t]
    \centering
    \includegraphics[height=5.4cm]{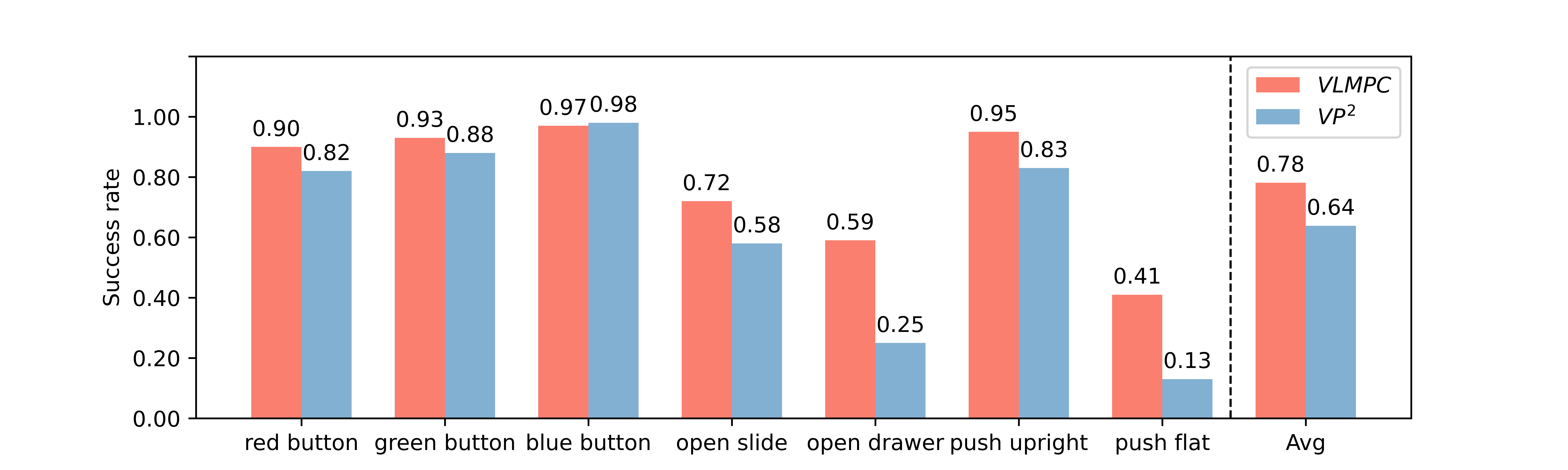}
    \captionsetup{justification=centerlast, singlelinecheck=false}
    \caption{Quantitative comparison with the $\text{VP}^2$ baseline in the RoboDesk environment.}
    \label{fig5}
\end{figure*}

\begin{figure}[t]
    \centering
    \includegraphics[width=1\linewidth]{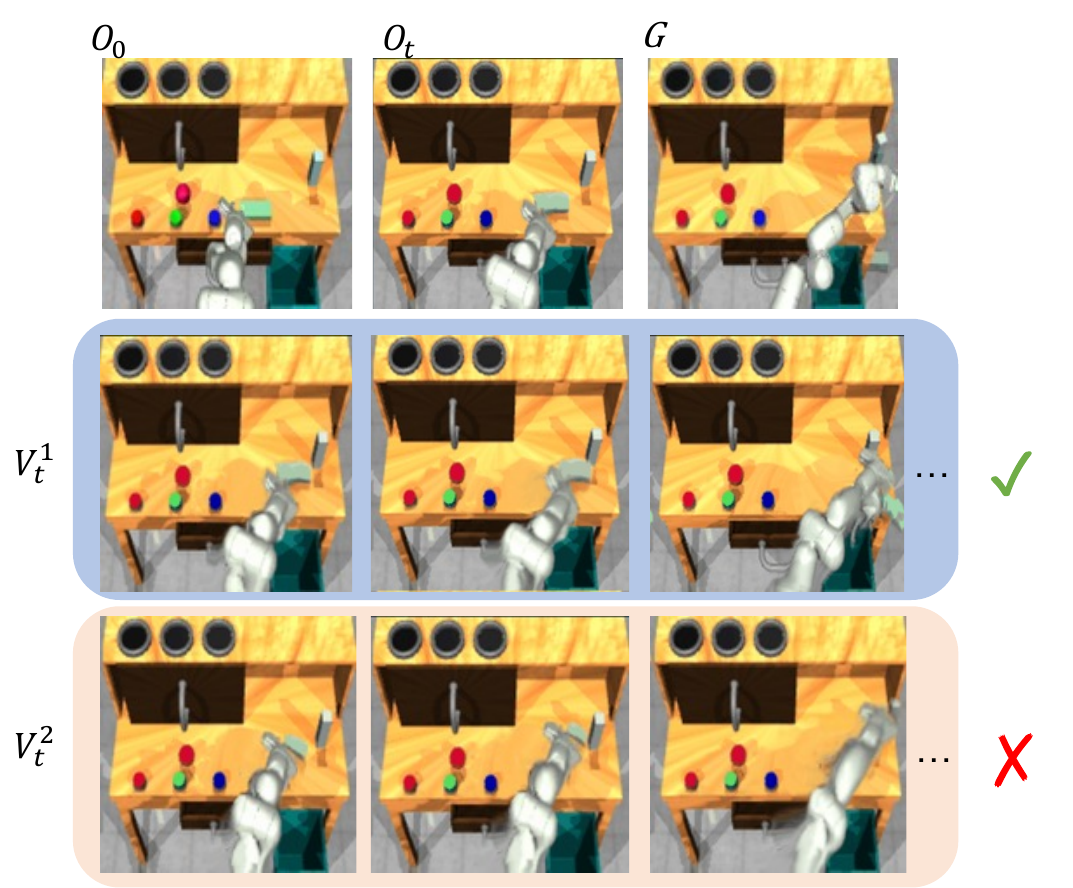}
    \caption{Visualization of the action-conditioned video predictions assessed by the hierarchical cost function of VLMPC via knowledge reasoning.}
    \label{fig6}
\end{figure}

\subsection{Simulation Experiments}

\subsubsection{Simulation Environments and Experimental Settings.}

The first evaluation is conducted on the popular simulation benchmark VP$^2$~\citep{vp2} which offers two environments RoboDesk~\citep{kannan2021robodesk} and \texttt{robosuite}~\citep{zhu2020robosuite}.
Considering the significant difference between the physical rendering of \texttt{robosuite} and real-world scenarios, we only use RoboDesk in this work. RoboDesk provides a physical environment with a Franka Panda robot arm, as well as 
a set of manipulation tasks. VP$^2$ conducts 7 sub-tasks: \emph{push \{red, green, blue\} button, open \{slide, drawer\}, push \{upright block, flat block\} off table.} For each sub-task, VP$^2$ provides 30 goal images as task input.

In the second experiment, we compare VLMPC with 5 existing methods in the Language Table environment~\citep{lynch2023interactive} on the \textit{move to area} task following VLP~\citep{du2024video}. Such a task is given by the language instruction: \textit{move all blocks to different areas of the board}. The 5 competing methods are UniPi~\citep{du2023learning}, LAVA~\citep{lin2023videollava}, PALM-E~\citep{palm-e}, RT-2~\citep{rt-2}, and VLP. We follow VLP to compute rewards using the ground truth state of each block in the Language Table environment. And we evaluate the methods on 50 randomly initialized environments.


\subsubsection{Experimental Results.}
The experimental results on the $\text{VP}^2$ benchmark are listed in Fig.~\ref{fig5}. It can be seen that VLMPC significantly outperforms the VP$^2$ baseline. We can see that for the tasks of \emph{push
\{red, green, blue\} button}, both the VP$^2$ baseline and VLMPC achieve high performance. This is simply because such tasks contain no multiple sub-goals. Thus, once the robot arm reaches the specific button and pushes it, the task is completed. On the other hand, the remaining tasks are more challenging, which require the robot to identify and move among multiple sub-goals as well as avoiding collision with interference objects. We can see that VLMPC significantly outperforms the VP$^2$ baseline in such challenging tasks, demonstrating its good reasoning and planning capability.

\begin{table}[t]
\setlength\tabcolsep{3pt}
\linespread{1.2}
\begin{center}\small
\caption{Comparison with existing methods on the task of \textit{move to area}  in the Language Table environment.}

\begin{tabular}{c|cc}
\hline\hline
Method & Success Rate(\%)  & Reward \\
\hline\hline
{UniPi~\citep{du2023learning}} & 0 & 30.8 \\
{LAVA~\citep{lin2023videollava}} & 22 & 59.8 \\
{PALM-E~\citep{palm-e}} & 0 & 36.5 \\
{RT-2~\citep{rt-2}} & 0 & 18.5 \\
{VLP~\citep{du2024video}} & 64 & 87.3 \\
{VLMPC} & \textbf{70} & \textbf{89.3} \\

\hline\hline
\end{tabular}
\label{tab_baseline}
\end{center} 
\end{table}

\begin{figure*}[t]
    \centering
\includegraphics[width=0.8\linewidth]{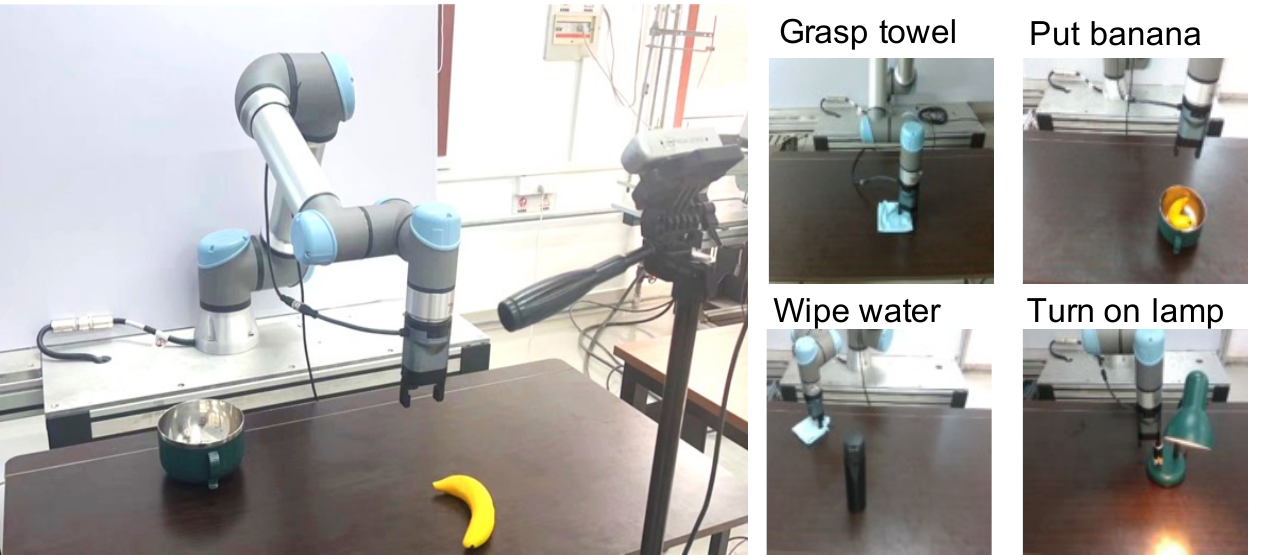}
    \caption{The real-world experimental platform includes a UR5 robot arm and a monocular RGB camera. It also shows a goal image for each of the four tasks.}
    \label{fig7}
\end{figure*}

Fig.~\ref{fig6} shows the visual results for the most challenging sub-task \emph{push flat}. This task requires pushing a flat green block off the table, while keeping other objects unmoved. We notice a slender block standing on the right edge of the table, which obviously serves an interference object. 
For the current observation $O_t$, we select two predicted videos for visualization. The second and the third rows of Fig.~\ref{fig6} show the predicted videos corresponding to different candidate action sequences. It can be seen that both candidate action sequences have the tendency to push the flat block off the table. It is noteworthy that the VP$^2$ baseline using a pixel-level cost and a simple state classifier assigns similar costs on both videos, which leads to the selection of an inappropriate action sequence. In contrast, VLMPC produces a higher cost for $V_t^2$ which contains a possible collision between the robot arm and the interference object. $V_t^1$ indicates a more reasonable moving direction and interaction with objects, and is thus assigned a lower cost. Such results demonstrate that the proposed hierarchical cost function can make the desired assessment of the predicted videos on the knowledge level and facilitate VLMPC to select an appropriate action to execute. 

Table~\ref{tab_baseline} lists the quantitative results of the comparative experiment conducted in the Language Table environment, where the Reward metric is computed in accordance with the VLP reward. It can be seen that our VLMPC outperforms all competing methods. This is because VLMs are good at localizing specific areas. Therefore, through sampling actions towards the sub-goals, VLMPC enables the robot to successfully reach the sub-goals and complete the task.





\subsection{Real-World Experiments}
\subsubsection{Experimental Setting.}
As shown in Fig.~\ref{fig7}, we use a UR5 robot to conduct real-world experiments. A monocular RGB camera is set up in front of the manipulation platform to provide the observations. We design four manipulation tasks, including \emph{grasp towel}, \emph{put banana}, \emph{turn on lamp}, and \emph{wipe water}. In each manipulation task, the position of the objects is initialized randomly within the reachable space of the action, yielding different goal images.  It is noteworthy that the objects involved in these tasks are not included in the collected data for training the video prediction model.

\begin{table}[t]
\setlength\tabcolsep{1pt}
\linespread{1.2}
\begin{center}\small
\caption{Results of VLMPC using goal image or language instruction as input in real-world experiments.}
\begin{tabular}{c|cc|cc}
\hline\hline
\multirow{2}{*}{Tasks} & \multicolumn{2}{c|}{Goal Image} & \multicolumn{2}{c}{Language Instruction} \\
 & Success Rate(\%)  & Time(s) & Success Rate(\%)  & Time(s) \\
\hline\hline
\emph{grasp towel} & 76.67 & 162.4 & 73.33 & 184.6 \\
\emph{put banana} & 60.00 & 203.9 & 46.67 & 230.7\\
\emph{turn on lamp} & 83.33 & 128.4 & 86.67 & 142.8\\
\emph{wipe water} & 36.67 & 289.3 & 23.33 & 331.9\\

\hline\hline
\end{tabular}
\label{tab1}
\end{center} 
\end{table}

\begin{table*}[ht]
\setlength\tabcolsep{1pt}
\linespread{1.2}
\begin{center}\small
\caption{Ablation study using the variants of VLMPC on different tasks in real-world environments.}

\begin{tabular}{c|cc|cc|cc|cc}
\hline\hline
\multirow{2}{*}{VLMPC Variant} & \multicolumn{2}{c|}{\emph{grasp towel}} & \multicolumn{2}{c|}{\emph{put banana}} & \multicolumn{2}{c|}{\emph{turn on lamp}} & \multicolumn{2}{c}{\emph{wipe water}}\\
 & Success Rate(\%)  & Time(s) & Success Rate(\%)  & Time(s) & Success Rate(\%)  & Time(s) & Success Rate(\%)  & Time(s)\\
\hline\hline
{VLMPC-RS}  & 63.33 & 302.5 & 40 & 389.5 & 73.33 & 256.7 & 13.33 & 573.9\\
{VLMPC-PD} & 26.67 & 178.3 & 0 & - & 60.00 & \textbf{123.6} & 0 & -\\
{VLMPC-VS} & 56.67 & 201.5 & 46.67 & 297.3 & 56.67 & 243.7 & 10.00 & 543.9\\
{VLMPC-MCVD} & 33.33 & 509.3 & 23.33 & 689.4 & 46.67 & 553.8 & 6.67 & 803.5\\
{VLMPC} & \textbf{76.67} & \textbf{162.4} & \textbf{60.00} & \textbf{203.9} & \textbf{83.33} & 128.4 & \textbf{36.67} & \textbf{289.3}\\

\hline\hline
\end{tabular}
\label{tab_ablation}
\end{center} 
\end{table*}

\begin{table}[t]
\setlength\tabcolsep{1pt}
\linespread{1.2}
\begin{center}\small
\caption{Comparison between Traj-VLMPC and VLMPC using language instruction as task input in real-world experiments.}
\begin{tabular}{c|cc|cc}
\hline\hline
\multirow{2}{*}{Tasks} & \multicolumn{2}{c|}{VLMPC} & \multicolumn{2}{c}{Traj-VLMPC} \\
 & Success Rate(\%)  & Time(s) & Success Rate(\%)  & Time(s) \\
\hline\hline
\emph{grasp towel} & 73.33 & 184.6 & 93.33 & 68.1 \\
\emph{put banana} & 46.67 & 230.7 & 80.00 & 105.9 \\
\emph{turn on lamp} & 86.67 & 142.8 & 83.33 & 54.0\\
\emph{wipe water} & 23.33 & 331.9 & 66.67 & 126.6\\

\hline\hline
\end{tabular}
\label{tab5}
\end{center} 
\end{table}

\subsubsection{Experimental Results.}
To properly evaluate VLMPC in real-world tasks, we repeat each task 30 times by randomly initializing the position of all objects and change the color of the tablecloth every 10 times. We calculate the success rate and the average time for each task respectively. The results are listed in Table~\ref{tab1}. It can be seen that VLMPC achieves high success rates for the tasks of \emph{grasp towel} and \emph{turn on lamp}. The two tasks are relatively simple as there is no interference object in the scene. The success rates for the tasks of \emph{put banana} and \emph{wipe water} are low as they are more challenging. \emph{put banana} contains multiple sub-goals, and \emph{wipe water} is even more difficult as it involves both interference objects and multiple sub-goals. Such results demonstrate that VLMPC generalizes well to novel objects and scenes unseen in the training dataset.

\begin{figure*}[t]
    \centering
\includegraphics[width=1\linewidth]{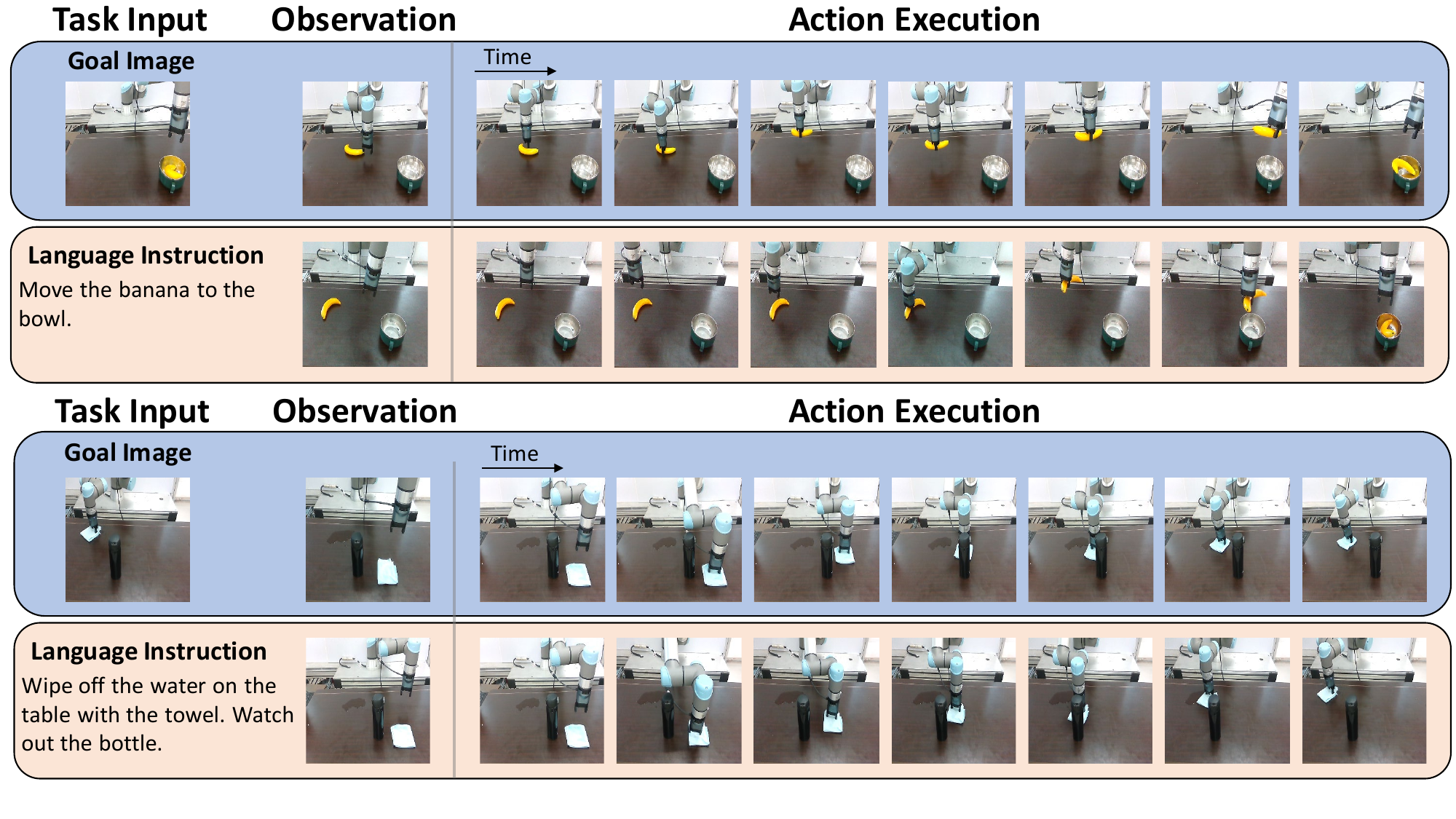}
    \caption{Action execution in VLMPC for two challenging real-world manipulation tasks \emph{put the banana in the bowl} and \emph{wipe water}.}
    \label{fig8}
\end{figure*}

We also provide the visual results for two challenging tasks \emph{put banana in the bowl} and \emph{wipe water}. As shown in Fig.~\ref{fig8}, in the \emph{put banana in the bowl} task, VLMPC correctly identifies the first sub-goal, \ie the banana, based on the current observation, and drives the robot arm moving towards and finally grasping it. Then, VLMPC dynamically finds the next sub-goal, \ie the bowl, and subsequently guides the robot to move to the area above it and opens the gripper. This example demonstrates that VLMPC has the desired capability of dynamically identifying the sub-goals during the task. The \emph{wipe water} task requires the robot arm to wipe off the water on the
table with the towel while watching
out the bottle. It is clear that this task contains two sub-goals \emph{towel} and \emph{water}, and an interference object \emph{bottle}. Fig.~\ref{fig8} shows that VLMPC successfully identifies all of them, and guides the robot to select appropriate actions to execute while avoiding the collision with the interference object. We provide more visualized results on four sub-tasks with both successful and failure cases, as well as related discussion in the supplementary material. We also provide video demonstrations in both simulated and real-world environments.

To evaluate the performance of Traj-VLMPC in real-world scenes, we conducted the same four tasks using language instructions. The results are listed in Table~\ref{tab5}. It can be seen that Traj-VLMPC outperforms VLMPC on all four tasks. In particular, even in the two challenging tasks of \emph{put banana} and \emph{wipe water}, which involve multiple sub-goals and interference objects, Traj-VLMPC still achieves a significantly higher success rate. This improvement can be attributed to VLMPC's inability to effectively avoid collision with interference objects when the end-effector is close to them, whereas Traj-VLMPC uses its GMM sampling module to generate trajectories that bypass the obstacles.

Furthermore, the task completion time of Traj-VLMPC is significantly shorter than that of VLMPC. This is primarily due to Traj-VLMPC's ability to directly sample long-horizon trajectories and execute the optimal trajectory selected by the cost function, enabling more efficient task execution. In contrast, VLMPC adopts a step-by-step framework that requires evaluating a batch of video predictions at each time step through VLMs, leading to increased time consumption. These results highlight Traj-VLMPC's superior trajectory sampling capability for collision avoidance and its enhanced long-horizon planning capability, which collectively reduce time consumption and improve overall performance.

\subsubsection{Applying Traj-VLMPC in Long-Horizon Tasks.} 
\begin{figure*}[t]
    \centering
\includegraphics[width=1\linewidth]{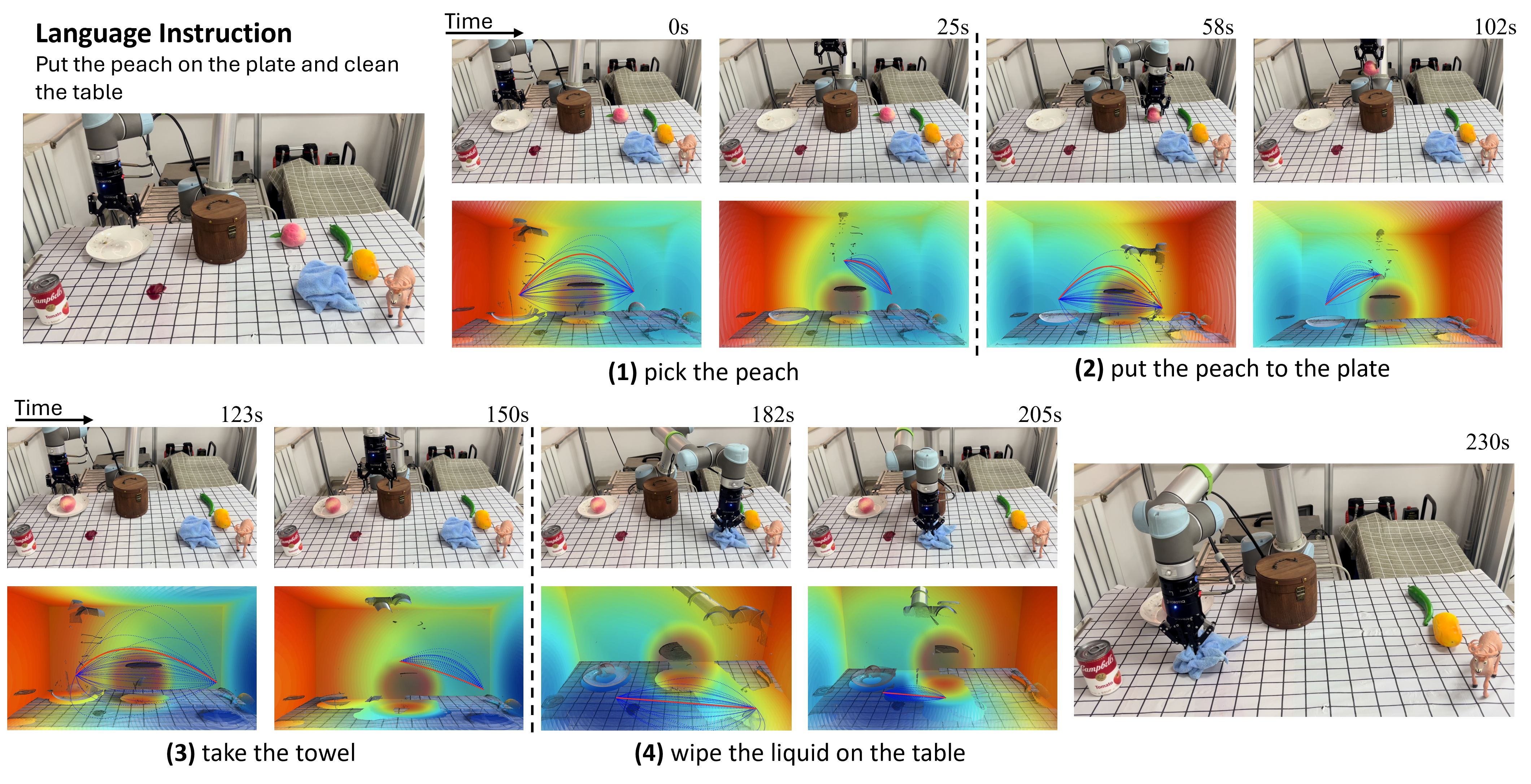}
    \caption{Robotic manipulation with Traj-VLMPC in a long-horizon real-world task following the instruction \emph{``Put the peach on the plate and clean the table.''}. VLM first decomposes the command into four sub-tasks, and then each sub-task is executed with real-time trajectory sampling and the corresponding voxel-based 3D value map, ensuring efficient and collision-free execution.}
    \label{fig_trajvlmpc_long_task}
\end{figure*}

By substantially improving the efficiency and reducing the computational overhead of VLMPC, Traj-VLMPC enables real-time robotic manipulation and can be applied to more complex, long-horizon tasks. In such tasks, VLM can easily decompose the overall goal into multiple sub-tasks. For instance, as shown in Fig.~\ref{fig_trajvlmpc_long_task}, the task ``put the peach on the plate and clean the table'' can be split into sub-tasks: (1) \emph{pick the peach}, (2) \emph{push the peach onto the plate,} (3) \emph{take the towel,} and (4) \emph{wipe the liquid on the table.} For the sub-goal of each sub-task, Traj-VLMPC repeatedly constructs a GMM to generate trajectory candidates and evaluates them using a voxel-based 3D value map. This allows for rapid, real-time switching and execution across sub-tasks. Compared to VLMPC which relies on step-by-step video prediction, Traj-VLMPC completes the entire task more reliably and faster, demonstrating its capability of real-time performance in long-horizon scenarios.

\subsection{Ablation studies}

We conducted ablation studies to demonstrate the effectiveness of each core component of VLMPC. In the experiments, we compare VLMPC with 4 variants described as follows:

\textbf{VLMPC-RS}: This is an ablated version of VLMPC where the conditional action sampling module is replaced with random sampling which simply sets the sampling mean $\mu_t$ to zero.

\textbf{VLMPC-PD}: This variant of VLMPC only uses the pixel distance cost as the cost function.

\textbf{VLMPC-VS}: This variant of VLMPC only uses the VLM-assisted cost as the cost function.

\textbf{VLMPC-MCVD}: In this variant of VLMPC, we replace DMVFN-Act with the action-conditioned video prediction model MCVD \citep{vp2,voleti2022mcvd}.

Table~\ref{tab_ablation} lists the results. First, compared with random sampling, our conditional action sampling module makes the robot complete various tasks more quickly and achieve higher success rates. This is because random sampling cannot make the sampled action sequences focus on the direction to sub-goals. Second, when VLMPC only uses the pixel distance cost, we found that the robot directly moves to the goal position and ignores intermediate sub-goals, leading to low success rates in the tasks \textit{put banana} and \textit{wipe water}. Besides, when VLMPC only uses the VLM-assisted cost, we found that VLM sometimes localizes incorrect sub-goals, which also leads to low success rates. Third, compared with DMVFN-Act, the diffusion-based video prediction model MCVD leads to much lower efficiency in all testing tasks.

\section{Conclusion}
\label{sec:conclusion}
This paper introduces VLMPC that integrates VLM with MPC for robotic manipulation. It prompts VLM to produce a set of candidate action sequences conditioned on the knowledge reasoning of goal and observation, and then follows the MPC paradigm to select the optimal one from them. The hierarchical cost function based on VLM is also designed to provide an amenable assessment for the actions through estimating future frames generated by a lightweight action-conditioned video prediction model. Experimental results demonstrate that VLMPC performs well in both simulated and real-world scenarios. 

\textbf{Limitation.} VLMPC faces a significant limitation as it relies on step-by-step video prediction, leading to high computational costs and difficulties when handling longer-horizon tasks. Traj-VLMPC addresses this issue by using trajectory-based sampling and assessment to reduce the frequency of per-step VLM queries. However, to maintain efficiency, Traj-VLMPC avoids video prediction, which may reduce the robustness in unexpected situations and restrict the adaptability in dynamic environments. Hence, integrating a more powerful and efficient video prediction model (\textit{e.g.} an advanced world model) to provide real-time and reliable prediction on the future state and designing a more efficient scheme for integrating VLM with MPC are of interest in future work.

\section*{Acknowledgment}

This work was supported in part by the National Natural Science Foundation of China under Grant U22A2057, in part by the Shandong Excellent Young Scientists Fund Program (Overseas) under Grant 2022HWYQ-042, and in part by the National Science and Technology Major Project of China under Grant 2021ZD0112002.





\bibliographystyle{SageH}
\bibliography{references}
\end{document}